\documentclass[runningheads]{llncs}

\usepackage{cite}
\usepackage{amsmath,amssymb,amsfonts}

\usepackage{booktabs}

\usepackage{siunitx}
\DeclareSIUnit{\nothing}{\relax}
\usepackage{algpseudocode,algorithm}
\usepackage{xcolor}
\usepackage{array}

\usepackage{graphicx}
\usepackage{grffile} 
\usepackage{float}
\usepackage{makecell}
\usepackage{adjustbox} 

\usepackage{etoolbox} 

\usepackage{url}
\def\MYTITLE{Event-based Stereo Depth Estimation from Ego-motion using Ray Density Fusion}

\newcommand\MYhyperrefoptions{bookmarks=true,bookmarksnumbered=true,
pdfpagemode={UseOutlines},plainpages=false,pdfpagelabels=true,
colorlinks=true,breaklinks=true,
pdftitle={\MYTITLE},%
pdfsubject={Computer Vision, Depth Estimation, Robotics},%
pdfauthor={S. Ghosh, G. Gallego},%
pdfkeywords={Event camera, Asynchronous sensor, 3D Reconstruction, Space Sweep, Stereo}}%
\usepackage[\MYhyperrefoptions,pdftex]{hyperref}

\usepackage{orcidlink}

\usepackage[capitalize]{cleveref}
\newif\ifaisy
\aisyfalse 
\ifaisy
\crefname{section}{Section}{Sections}
\crefname{table}{Table}{Tables}
\crefname{figure}{Figure}{Figures}
\else 
\crefname{section}{Sec.}{Secs.}
\crefname{table}{Tab.}{Tabs.} 
\crefname{figure}{Fig.}{Figs.}
\fi
\Crefname{section}{Section}{Sections}
\Crefname{table}{Table}{Tables}
\Crefname{figure}{Figure}{Figures}

\pdfpxdimen=\dimexpr 1 in/72\relax
\usepackage{calc}

\newif\ifclearsectionlook
\clearsectionlookfalse

\def\Lum{L}
\def\pol{p} 

\def\bx{\mathbf{x}}

\def\bX{\mathbf{X}}

\robustify \bfseries
\robustify \underline

\definecolor{light-gray}{gray}{0.75}
\newcommand\gframe[1]{\color{light-gray}\frame{#1}}

\usepackage[absolute]{textpos}

\begin{document}

\definecolor{somegray}{gray}{0.6}
\newcommand{\darkgrayed}[1]{\textcolor{somegray}{#1}}
\begin{textblock}{8}(4, 0.8)
\begin{center}
\darkgrayed{This extended abstract has been accepted at the 2nd International \\ 
Ego4D Workshop of the European Conf.~on Computer Vision, 2022.}
\end{center}
\end{textblock}

\pagestyle{headings}
\mainmatter

\title{\MYTITLE}

\titlerunning{Event-based Stereo Depth Estimation from Ego-motion}
\authorrunning{S. Ghosh and G. Gallego}

\author{Suman Ghosh\inst{1}\orcidlink{0000-0002-4297-7544} 
\and Guillermo Gallego\inst{1,2}\orcidlink{0000-0002-2672-9241}
}

\institute{%
Department of EECS, Technische Universit{\"a}t Berlin, Berlin, Germany.  
\and
Einstein Center Digital Future and SCIoI
Excellence Cluster, Berlin, Germany. 
\vspace{-1.5ex}
}

\maketitle

\begin{abstract}
Event cameras are bio-inspired sensors that mimic the human retina by responding to brightness changes in the scene. 
They generate asynchronous spike-based outputs at microsecond resolution, providing advantages over traditional cameras like high dynamic range, low motion blur and power efficiency. 
Most event-based stereo methods attempt to exploit the high temporal resolution of the camera and the simultaneity of events across cameras to establish matches and estimate depth.
By contrast, this work investigates how to estimate depth from stereo event cameras without explicit data association by fusing back-projected ray densities, and demonstrates its effectiveness on head-mounted camera data, which is recorded in an egocentric fashion.
Code and video are available at \url{https://github.com/tub-rip/dvs_mcemvs}
\end{abstract}
\section{Introduction}
\label{sec:intro}

Event cameras, such as the Dynamic Vision Sensor \cite{Lichtsteiner08ssc,Finateu20isscc} (DVS), mimic the transient visual pathway in humans.
In contrast to traditional frame-based cameras, they sample the scene asynchronously, producing a stream of spikes, called ``events'', that encode the time, location and sign of per-pixel brightness changes. 
Events are naturally produced by ego-motion or moving objects in the scene.
Studying how the motion of a stereo event camera affects 3D visual perception could improve understanding of binocular vision and help design algorithms as efficient and robust as biological systems.
Moreover, the outstanding properties of event cameras, such as high dynamic range (HDR), high temporal resolution ($\approx$ \si{\micro\second}) and low power consumption, 
offer potential to tackle scenarios that are challenging for standard cameras (high speed and/or HDR) \cite{Gallego20pami,Delbruck13fns,Amir17cvpr,Gallego18cvpr,Rebecq18ijcv,Plizzari22cvpr,Shiba22eccv}.

This extended abstract is based on our recent paper on multi-event camera depth estimation \cite{Ghosh22aisy}. 
Similar to how humans exploit ego-motion of their eyes to explore and perceive the 3D scene, we use the motion of stereo event cameras to estimate scene depth (i.e., 3D reconstruction).

Most event-based stereo depth estimation methods use the epipolar constraint and the assumption of \emph{temporal coincidence} of events across retinas, i.e., a moving object produces events of same timestamps on both cameras \cite{Piatkowska13iccvw,Firouzi16npl,Osswald17srep,Carneiro13nn,Martel18iscas,Ieng18fnins}.
This relies on the high temporal resolution and sparse non-redundant output of event cameras to establish event matches across image planes and then triangulate the location of the 3D points.
\begin{figure*}[t]
\centering
    {
    \includegraphics[width=0.2\linewidth]{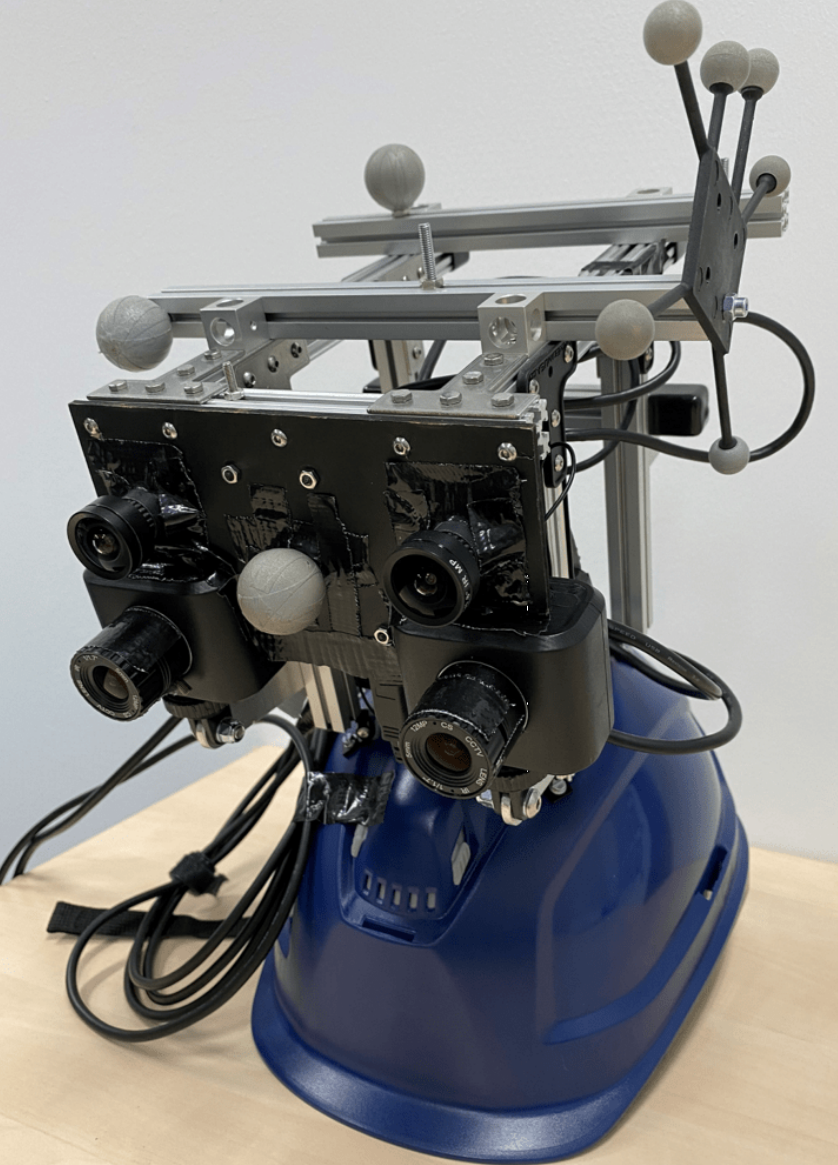}
    \hspace{0.1cm}
    \includegraphics[trim={6cm 11.2cm 0 0.7cm},clip,width=0.72\linewidth]{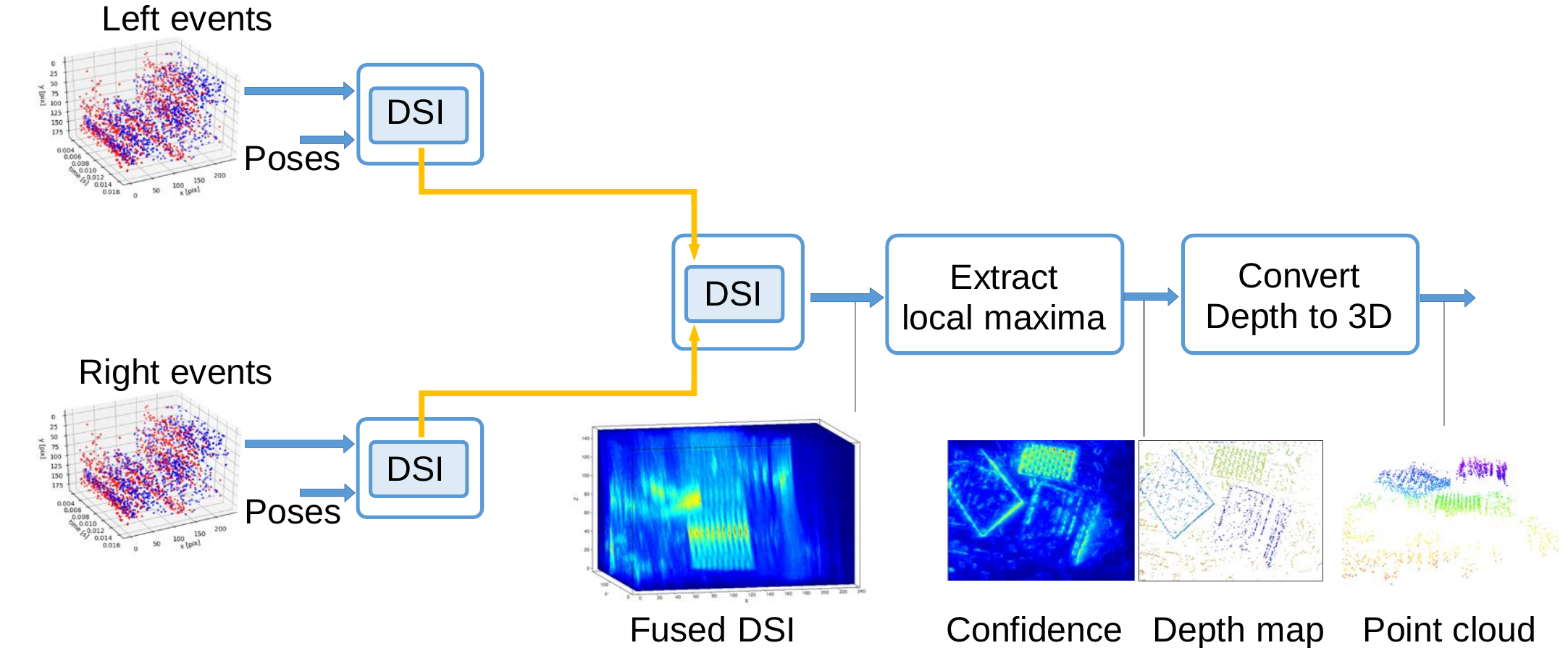}}
    \caption{Left: Head-mounted camera setup from the TUM-VIE dataset. 
    Image adapted from \cite{Klenk21iros}. 
    Right: Our event processing pipeline which uses the events from two or more synchronized, rigidly attached event cameras and their poses to estimate the scene depth.
    Using Space Sweeping, it builds ray density Disparity Space Images (DSIs) from each camera data and fuses them into a single DSI, from which depth is recovered. 
    Image adapted with permission from \cite{Ghosh22aisy}.
    \vspace{-3ex}
    }
    \label{fig:block-diagram}
\end{figure*}

In contrast, in \cite{Ghosh22aisy} we investigate a novel stereo approach, without explicitly establishing event matches (e.g., via event coincidence). 
Correspondence-free depth estimation for monocular event cameras has been shown to generate state-of-the-art results in visual odometry \cite{Rebecq17ral}.
We extend this idea to the multi-camera setting (i.e., $\geq 2$ synchronized event cameras rigidly attached) by directly fusing back-projected rays across cameras.

\section{Event-based Stereo Depth Estimation}
\label{sec:method:stereodepth}

In each camera, an event $e_k \doteq (\bx_k, t_k, \pol_{k})$ is triggered if the logarithmic brightness $\Lum$ at a pixel exceeds a contrast sensitivity $\theta>0$, 
where $\bx_k\doteq (x_k, y_k)^{\top}$, $t_k$ (in \si{\micro\second}) and $\pol_{k} \in \{+1,-1\}$
are the spatio-temporal coordinates and polarity of the brightness change, respectively.
Assuming constant illumination, the number of events produced per pixel is proportional to the amount of scene motion and texture.
As the camera moves, events are triggered at an almost continuous set of viewpoints. 
Similar to the monocular method in \cite{Rebecq17ral}, known camera poses are used to back-project events into space in the form of rays, referred to as a Disparity Space Image (DSI). 
The local maxima of the DSI (where many rays meet) provide candidate locations for the 3D edges which produced the events.

In our multi-camera setup, we perform information fusion early in the processing pipeline: at the DSI stage (\cref{fig:block-diagram}).
However, instead of naively using the same DSI for ray counting from both cameras, we define separate DSIs for both cameras at the same viewpoint in space.
Having one DSI per camera allows us to preserve the origin of the event data, 
and having geometrically \emph{aligned} DSIs avoids resampling errors during fusion.
We then fuse the DSIs using element-wise operations like the Generalized means, which satisfy an order relationship \cite{Ghosh22aisy}:
\begin{equation}
    \label{eq:metrics:order}
\min \leq H \leq G \leq A \leq \operatorname{RMS} \leq \max.
\end{equation}
Given two ray densities defined on the same 3D volume, an ideal fusion function should emphasize the regions of high ray density on \emph{both} DSIs and inhibit the rest.
The geometric mean $G$, harmonic mean $H$ and $\min$ functions satisfy this ``AND" logic and are the best at removing depth outliers \cite[Tab.~1]{Ghosh22aisy}.

\begin{algorithm}[t!]
  \caption{Stereo event fusion across cameras}
  \label{alg:fusion:stereo}
  {\small
  \begin{algorithmic}[1]
    \State \emph{Input}: stereo events in interval $[0,T]$, camera poses and camera calibration.
    \State Define a reference view (RV) for both DSIs: the left camera pose at $t=T/2$.
    \State Create two DSIs by back-projecting events from each camera.
    \State Fusion: compute the pointwise harmonic mean of the DSIs.
    \State Extract the depth map $Z^\ast(x,y) \doteq \arg\max f(\bX(x,y))$ 
    and the confidence map $c^\ast(x,y) \doteq \max f(\bX(x,y))$ 
    from the fused DSI $f: \bX\in \text{Volume} \subset \mathbb{R}^3\to \mathbb{R}$.
  \end{algorithmic}
  }
\end{algorithm}

Figure~\ref{fig:block-diagram} depicts the block diagram of our method.
First, the aligned DSIs are populated with back-projected events from each camera (left/right), 
then they are \emph{fused} into a single one, 
and finally local maxima are extracted to produce semi-dense depth and confidence maps.
The steps are given in Alg.~\ref{alg:fusion:stereo}.

\section{Experiments on Stereo Ego-motion Dataset}
\label{sec:experim:tumvie}

\def\figWidth{0.234\linewidth}
\begin{figure*}[t]
	\centering
    {\small
    \setlength{\tabcolsep}{1pt}
	\begin{tabular}{
	>{\centering\arraybackslash}m{0.4cm} 
	>{\centering\arraybackslash}m{\figWidth} 
	>{\centering\arraybackslash}m{\figWidth}
	>{\centering\arraybackslash}m{\figWidth}
	>{\centering\arraybackslash}m{\figWidth}}
 	 
 	 & \multicolumn{2}{c}{With ground truth poses}
  	 & \multicolumn{2}{c}{Without ground truth poses}\\
 	 \cmidrule(l{1mm}r{1mm}){2-3} \cmidrule(l{1mm}r{1mm}){4-5}
        & \emph{6dof} & \emph{desk2} & \emph{skate-easy} & \emph{bike-easy}\\

		\rotatebox{90}{\makecell{Depth}}
		&\gframe{\includegraphics[width=\linewidth]{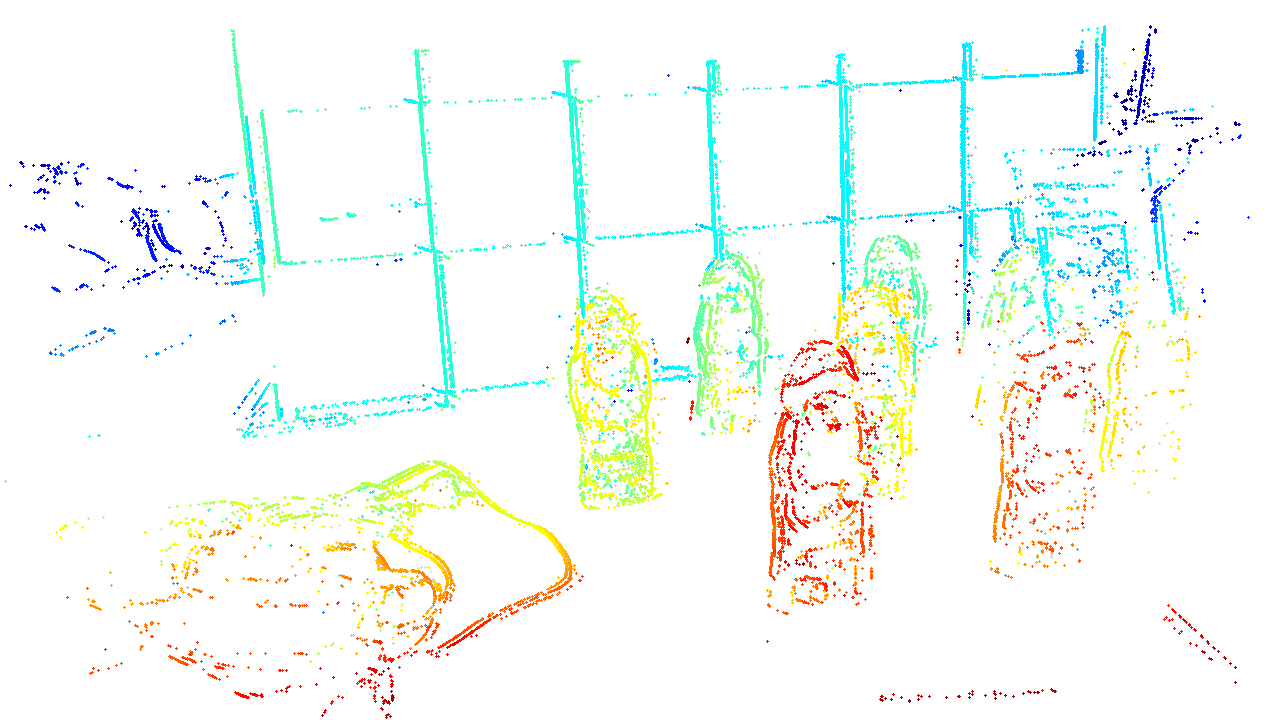}}
		&\gframe{\includegraphics[width=\linewidth]{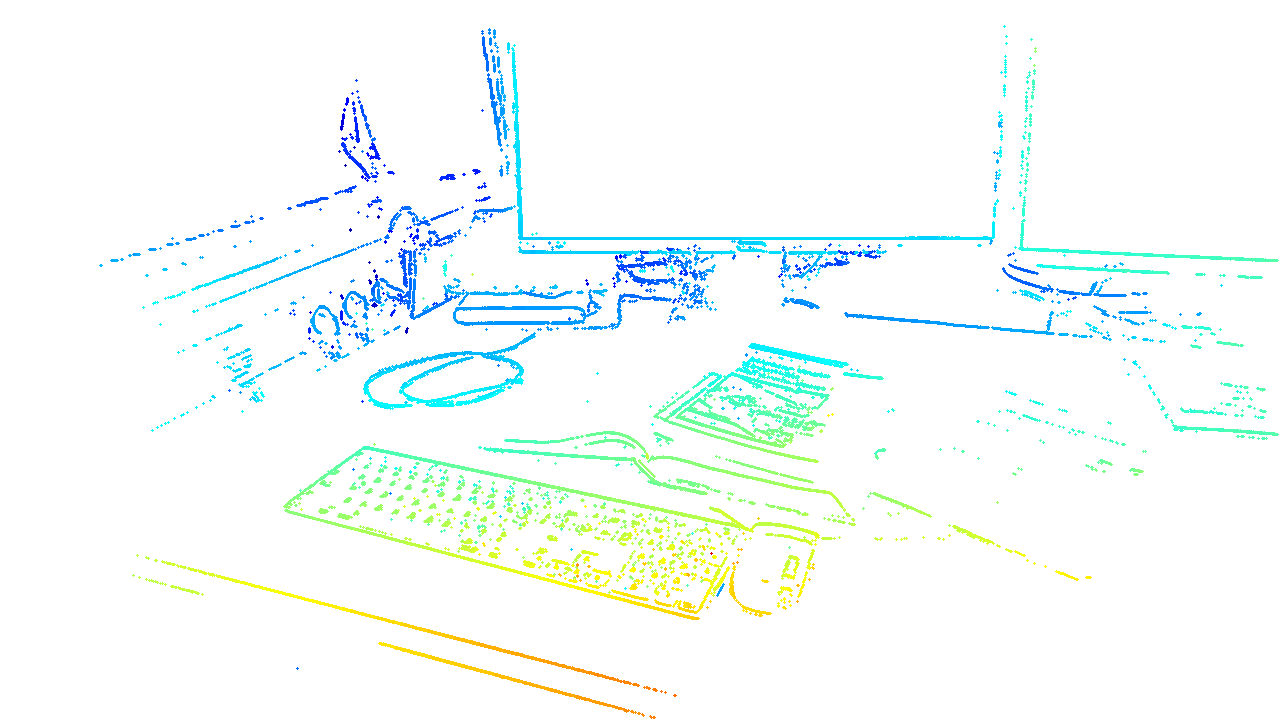}}
		&\gframe{\includegraphics[width=\linewidth]{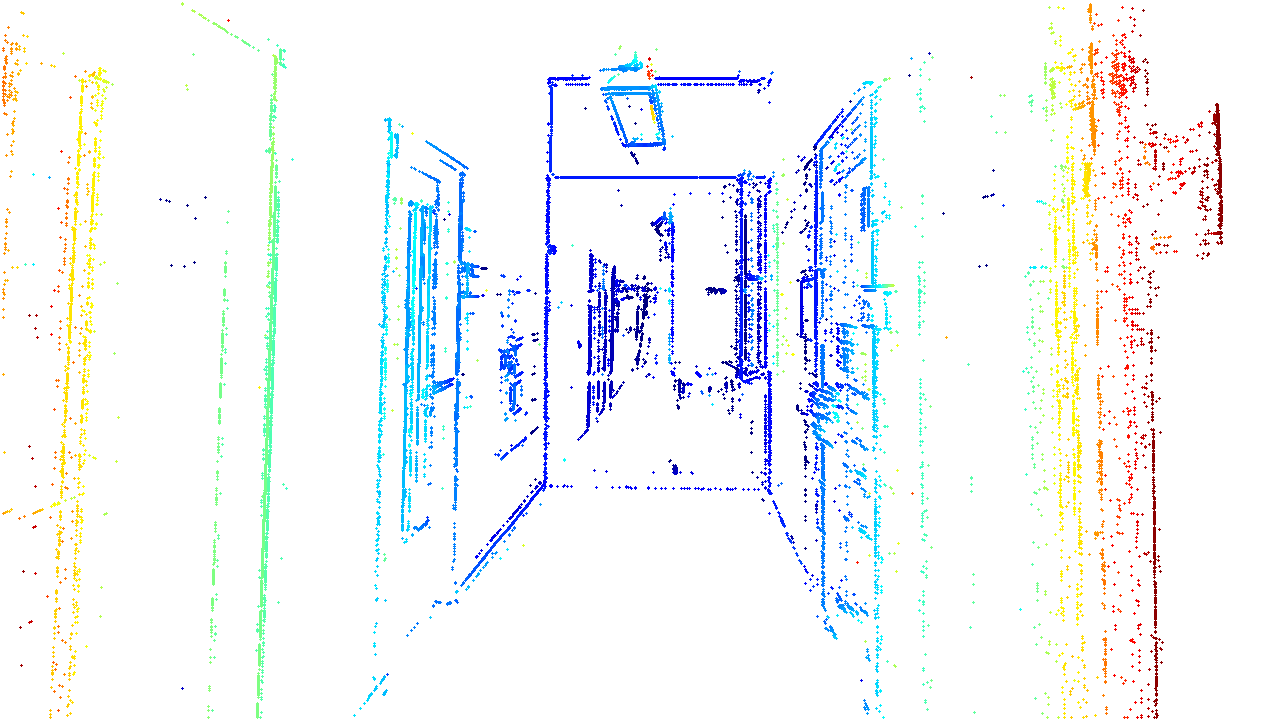}}
		&\gframe{\includegraphics[width=\linewidth]{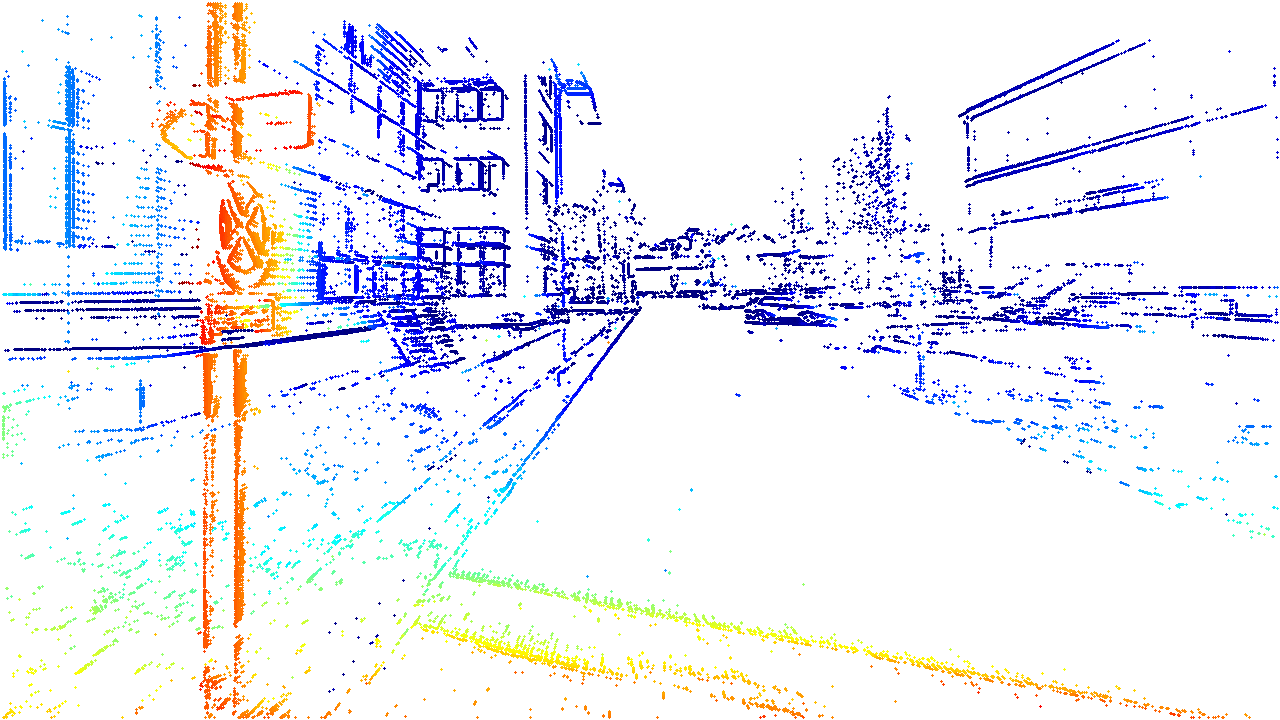}}
		\\
		
		\rotatebox{90}{\makecell{Confidence}}
		&\gframe{\includegraphics[width=\linewidth]{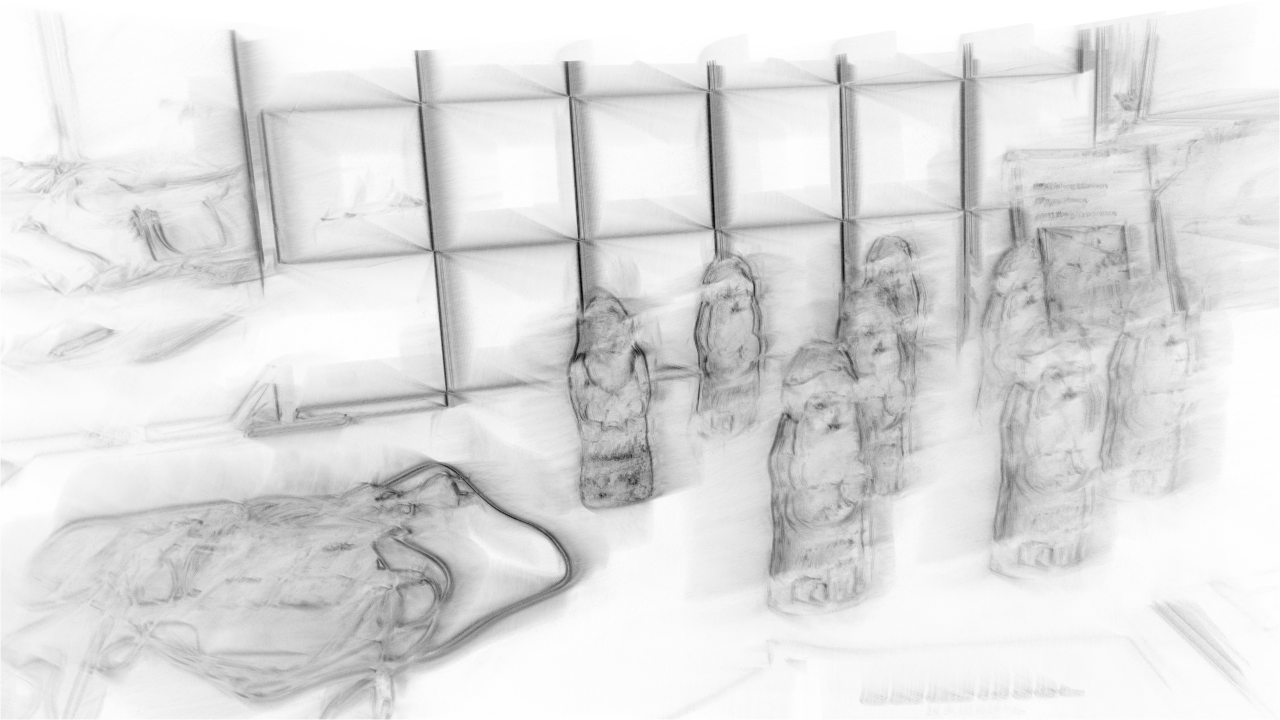}}
		&\gframe{\includegraphics[width=\linewidth]{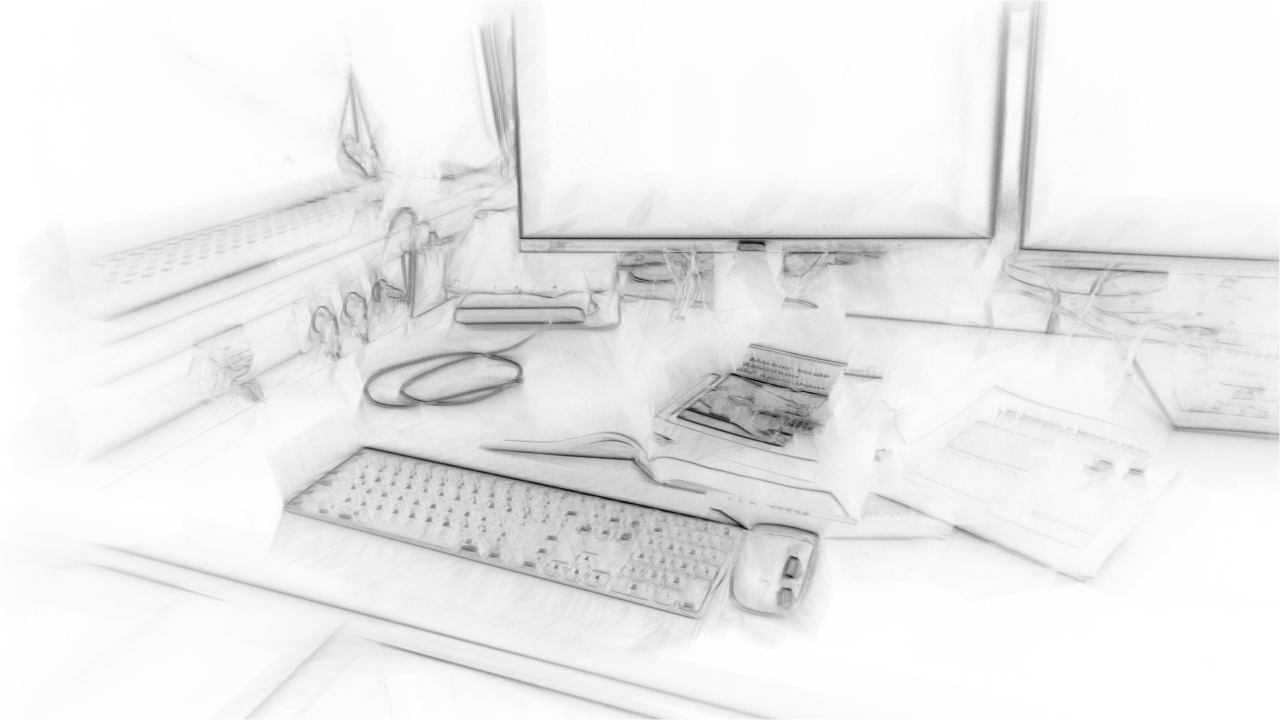}}
		&\gframe{\includegraphics[width=\linewidth]{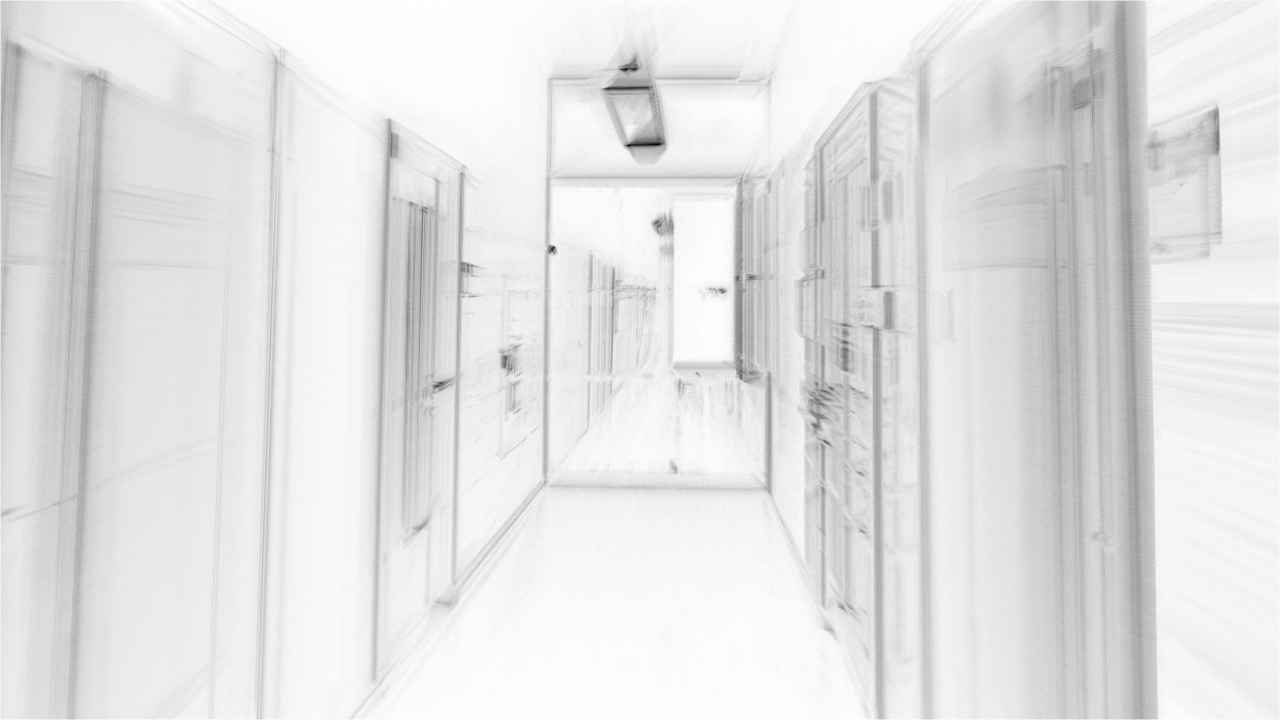}}
		&\gframe{\includegraphics[width=\linewidth]{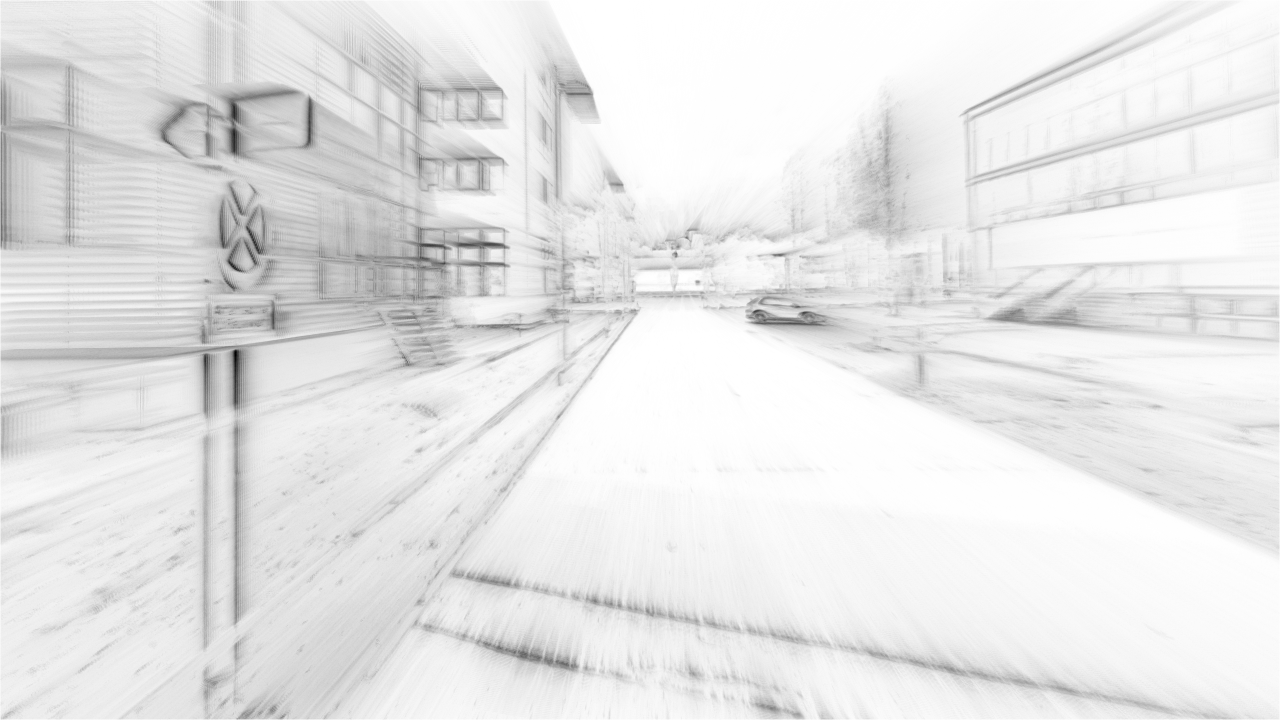}}		
		\\

	\end{tabular}
	}
 \vspace{-1ex}
	\caption{\label{fig:tumvie} Depth estimation using Alg.~\ref{alg:fusion:stereo} on TUM-VIE data \cite{Klenk21iros} (1Mpix cameras). 
    Depth is pseudo-colored from red (close) to blue (far), in the range 0.45--4 \si{\meter} (\emph{6dof} and \emph{desk2} sequences),
	1--20 \si{\meter} (\emph{skate-easy} corridor scene) 
    and 3--200 \si{\meter} (\emph{bike-easy} outdoors sequence).
    Reprinted with permission from \cite{Ghosh22aisy}.
 \vspace{-2ex}
	}
\end{figure*}

\def\figWidth{0.238\textwidth}
\begin{figure*}[t]
	\centering
    {\small
    \setlength{\tabcolsep}{2pt}
	\begin{tabular}{
	>{\centering\arraybackslash}m{\figWidth}
	>{\centering\arraybackslash}m{\figWidth}
	>{\centering\arraybackslash}m{\figWidth}
	>{\centering\arraybackslash}m{\figWidth}}
	     Frame (left camera) & Events (left camera) & Confidence map & Depth map\\
		\gframe{\includegraphics[trim={0px 130px 0 200px},clip,width=\linewidth]{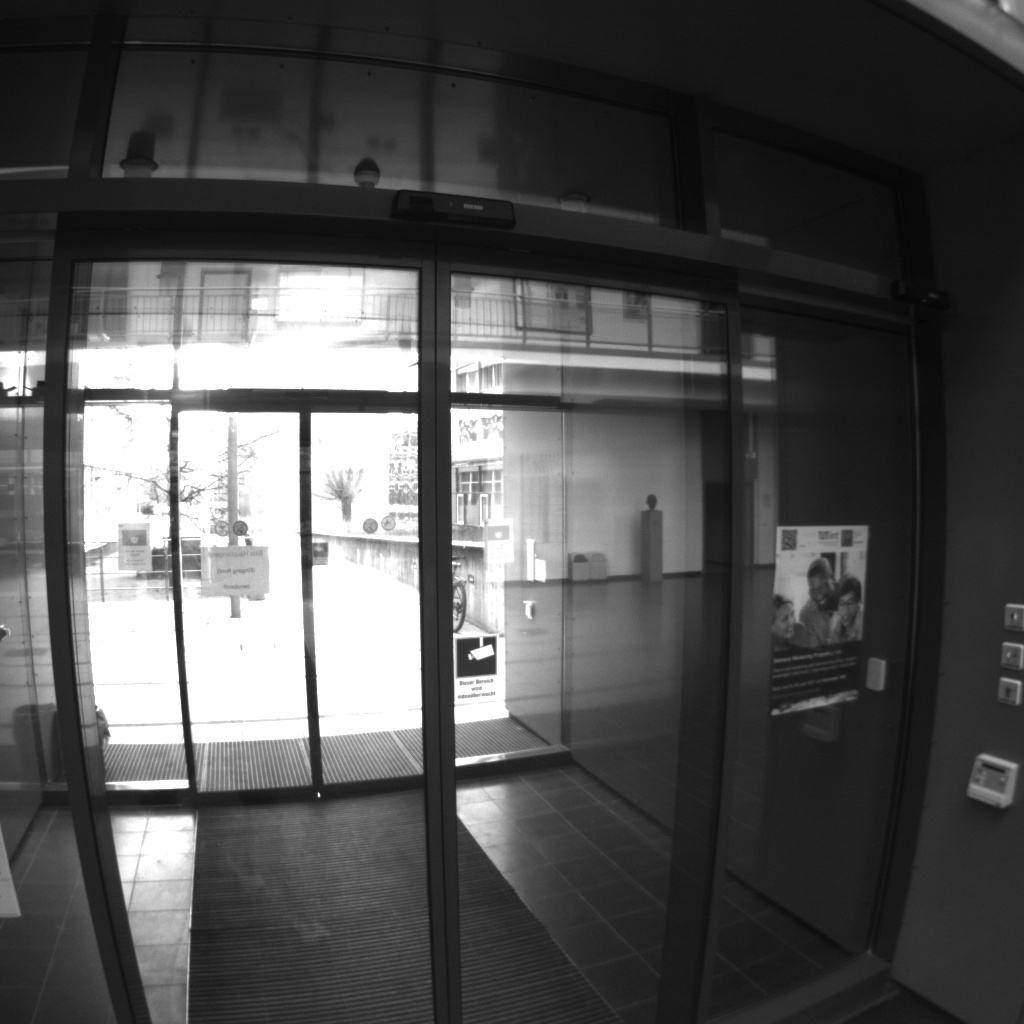}}
		&\gframe{\includegraphics[trim={0px 0 150px 0},clip,width=\linewidth]{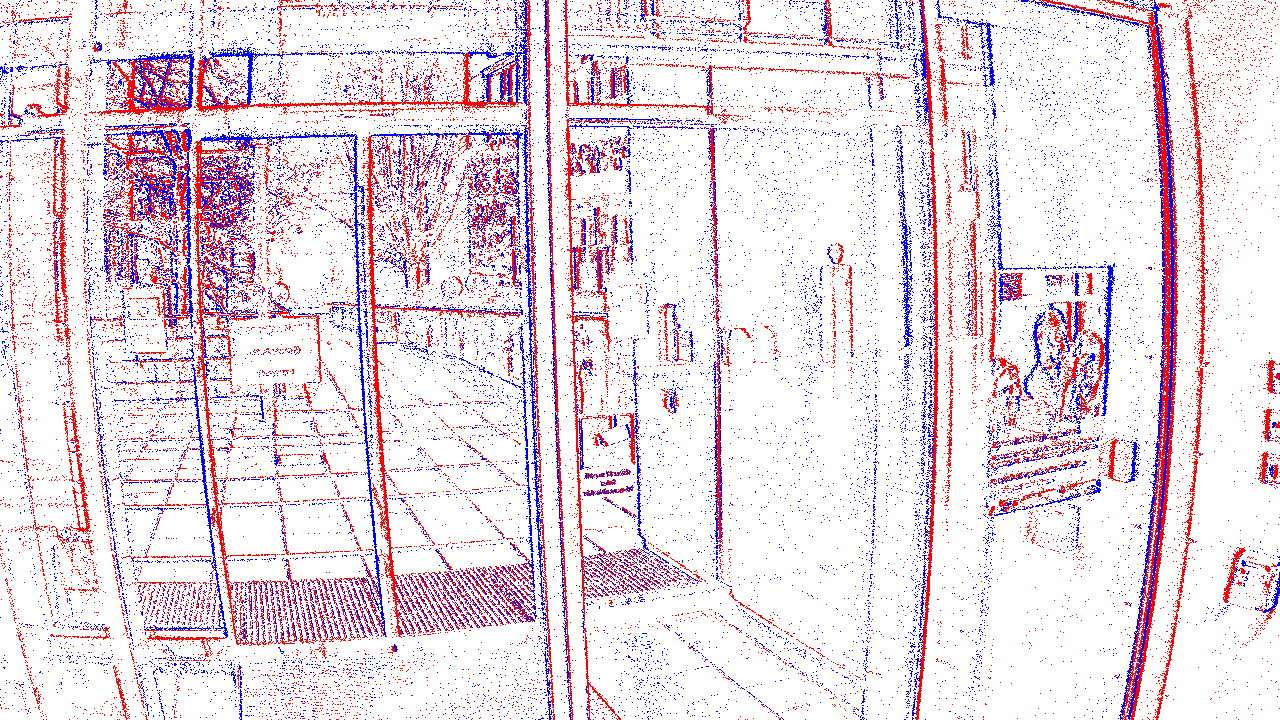}}
        &\gframe{\includegraphics[trim={0px 0 200px 0},clip,width=\linewidth]{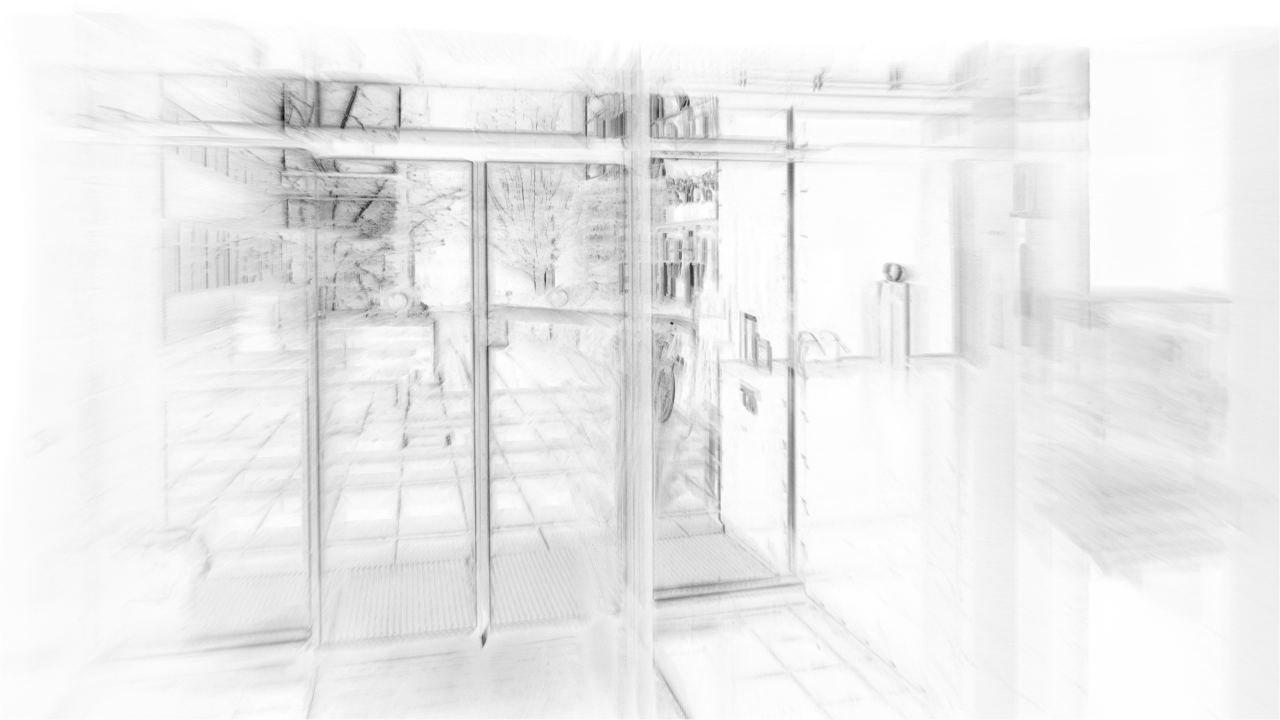}}
		&\gframe{\includegraphics[trim={0px 0 150px 0},clip,width=\linewidth]{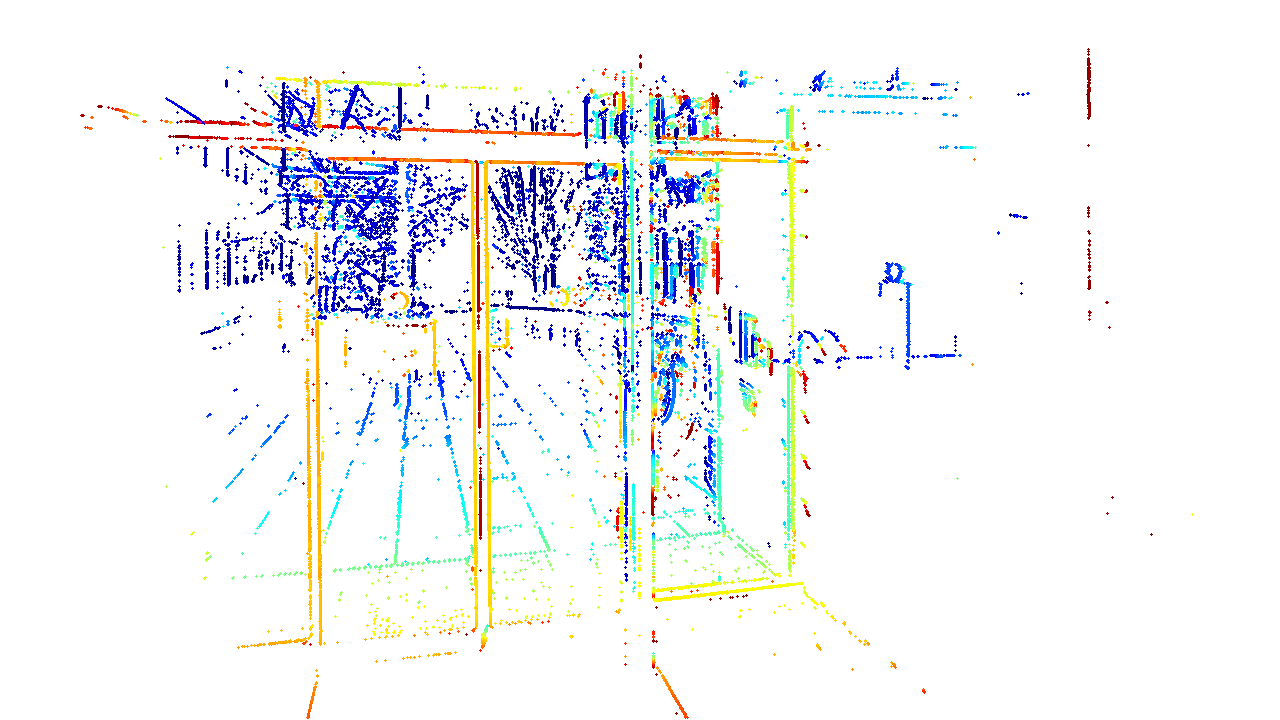}}
		\\
		
		\gframe{\includegraphics[trim={0px 150px 0 300px},clip,width=\linewidth]{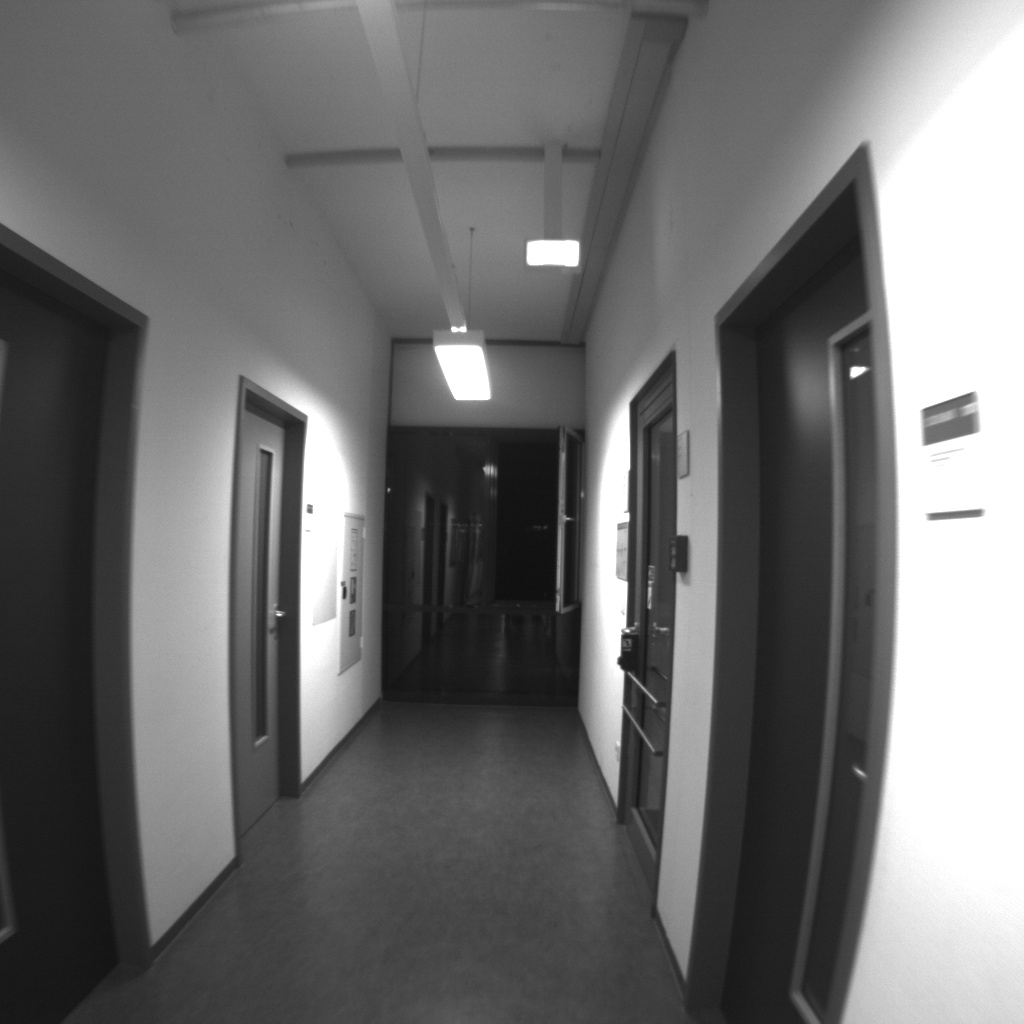}}
		&\gframe{\includegraphics[trim={0px 0 0 0},clip,width=\linewidth]{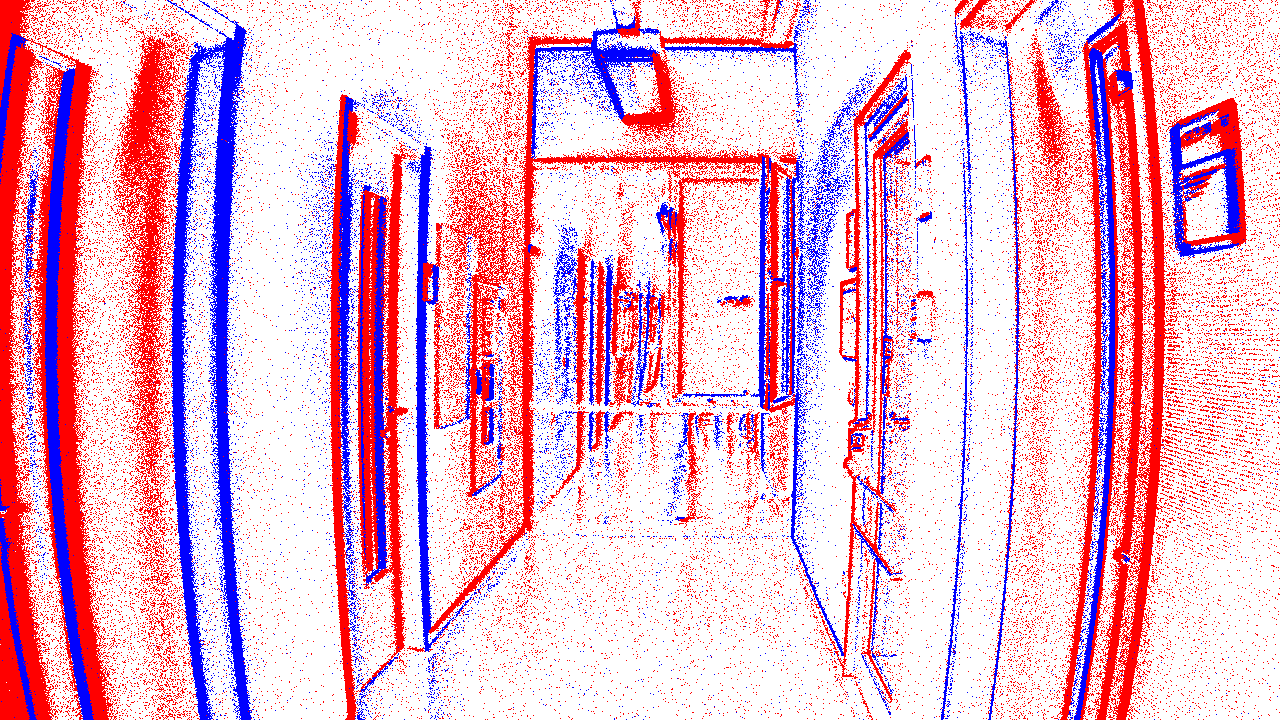}}
		&\gframe{\includegraphics[trim={0px 0 0 0},clip,width=\linewidth]{images/tumvie/skate-easy-1/15.000000confidence_map_fused_2.png}}
        &\gframe{\includegraphics[trim={0px 0 0 0},clip,width=\linewidth]{images/tumvie/skate-easy-1/15.000000inv_depth_colored_dilated_fused_2_w.png}}
		\\
	\end{tabular}
	}
	\caption{\label{fig:hdr}
 Output of Alg.~\ref{alg:fusion:stereo} on HDR scenes from the TUM-VIE dataset. 
 High dynamic range of event cameras allow our method to perceive both under- and over-exposed regions of the scene well.
 Reprinted with permission from \cite{Ghosh22aisy}. 
 \vspace{-3ex}
	}
\end{figure*}

While \cite{Ghosh22aisy} evaluates Alg.~\ref{alg:fusion:stereo} on a variety of datasets, here we present depth estimation results on the egocentric TUM-VIE dataset \cite{Klenk21iros} for the sake of brevity. 
Acquired using sensors mounted on a helmet, it contains sequences from a human's point-of-view while who performs locomotive tasks like walking, running, skating and biking, in various lighting conditions both indoors and outdoors. 
It is also the first public visual-inertial dataset with 1 Megapixel stereo event cameras \cite{Finateu20isscc}. 
To the best of our knowledge, we provide first results on this new event-based dataset.
We only present qualitative results because the dataset has no ground truth depth.

\Cref{fig:tumvie} presents results on indoor and outdoor sequences.
Despite the small depth range in the indoor scenes for the given camera baseline of (\SI{11.84}{\cm}),
our method can recover 3D structure, with clean depth maps containing few outliers. 
The space-sweeping method that builds the DSIs works best with lateral translations that produce parallax necessary for the convergence of the back-projected rays.
However, many sequences comprise forward camera motions, which contribute little parallax and produce fewer events; hence they are difficult for 3D reconstruction. 
The forward motion and the lower quality of camera poses in the last two columns of \cref{fig:tumvie} lead to an overall poorer reconstruction quality compared to the sequences recorded in the motion capture (mocap) room. 
In the accompanying video we provide a visual comparison between our method and state-of-the-art method ESVO \cite{Zhou20tro} on the TUM-VIE dataset.

Furthermore, we compared our method's performance with two different sources of ground truth poses (the mocap system and Basalt \cite{Usenko20ral}), 
and observed no major differences in the depth- and confidence maps.
We therefore concluded that ($i$) the poses from Basalt may be considered as accurate as the mocap for short time intervals (e.g., 0.5~\si{\second}, with $\approx$10M events), 
and ($ii$) our method is robust to noise: it estimates depth well using poses from a VIO (non-mocap) algorithm.

Additionally, we demonstrate the advantage of using events over standard frames for tackling HDR scenes in \cref{fig:hdr}. 
Our stereo method provides accurate depth estimates in all parts of the scene.

\vspace{-1ex}
\section{Conclusion}
\label{sec:conclusion}
\vspace{-1ex}
We presented a simple and effective algorithm for event-based multi-camera depth estimation.
With accurate camera poses, data association happens implicitly in DSI space, which removes the need for the event coincidence assumption.
The results show the effectiveness of our method on egocentric (head-mounted) stereo data, which resembles human binocular vision. 
We hope that our event-based depth estimator inspires the community to build robust and efficient systems for egocentric and robotic perception in challenging scenarios.

\textbf{Acknowledgements}.
Funded by the Deutsche Forschungsgemeinschaft (DFG, German Research Foundation) under Germany’s Excellence Strategy – EXC 2002/1 ``Science of Intelligence'' – project number 390523135.

\clearpage
%
\bibliographystyle{splncs04} 
\bibliography{all}
\end{document}